\newif\ifisdouble

\isdoubletrue 

\ifisdouble
	\documentclass{article}
	\usepackage{spconf}
\else
	\documentclass[11pt,draftcls,onecolumn]{IEEEtran}
\fi

\usepackage[utf8]{inputenc}

\usepackage[]{units}

\usepackage{subfigure}

\usepackage{amsmath, amsthm, url}
\usepackage{amssymb}
\usepackage{latexsym}

\usepackage{balance}

\usepackage{lipsum}

\usepackage[defblank]{paralist}

\usepackage[ruled]{algorithm2e}

\usepackage{dashrule}

\SetAlFnt{\small}
\SetAlCapFnt{\small}
\SetAlCapNameFnt{\small}
\SetAlCapHSkip{0pt}
\IncMargin{-\parindent}

\usepackage{ifpdf}

\ifpdf
\usepackage[pdftex]{graphicx}
\else
\usepackage{graphicx}
\fi

\ninept

\title{FAST KEYPOINT DETECTION IN VIDEO SEQUENCES}

\name{Luca Baroffio, Matteo Cesana, Alessandro Redondi, Marco Tagliasacchi\thanks{The project GreenEyes acknowledges the financial support of the Future and Emerging Technologies (FET) programme within the Seventh Framework Programme for Research of the European Commission, under FET-Open grant number:296676.}}
\address{Dipartimento di Elettronica, Informazione e Bioingegneria, Politecnico di Milano}

\begin{document}

\ifpdf
\DeclareGraphicsExtensions{.pdf, .jpg, .tif}
\else
\DeclareGraphicsExtensions{.eps, .jpg}
\fi

\newcommand{\x}{\mathbf{x}}
\newcommand{\xt}{\tilde{\x}}
\newcommand{\xh}{\hat{\x}}
\newcommand{\xb}{\bar{\x}}
\newcommand{\X}{\mathbf{X}}
\newcommand{\Xh}{\hat{\X}}
\newcommand{\Xt}{\tilde{\X}}

\newcommand{\Descr}{\mathcal{D}}
\newcommand{\Descrt}{\tilde{\Descr}}

\newcommand{\descr}{d}
\newcommand{\descrq}{\tilde{\descr}}

\newcommand{\dv}{\mathbf{\descr}}
\newcommand{\dq}{\tilde{\dv}}

\newcommand{\cv}{\mathbf{c}}
\newcommand{\cq}{\tilde{\cv}}

\newcommand{\Dim}{P}

\newcommand{\scale}{\sigma}
\newcommand{\scalet}{\tilde{\scale}}

\newcommand{\orient}{\theta}
\newcommand{\coord}{\mathbf{p}}
\newcommand{\coordq}{\tilde{\coord}}

\newcommand{\Rate}{R}
\newcommand{\Dist}{D}

\newcommand{\Image}{\mathcal{I}}

\newcommand{\T}{\mathbf{T}}

\newcommand{\Mask}{\mathcal{M}}
\newcommand{\Pyr}{\mathcal{L}}
\newcommand{\NOct}{\mathcal{O}}

\newcommand{\ThresID}{\mathcal{T}_{I}}
\newcommand{\ThresKH}{\mathcal{T}_{H}}
\newcommand{\ThresBM}{\mathcal{T}_{BM}}
\newcommand{\ThresET}{\mathcal{T}_{ET}}
\newcommand{\ThresRM}{\mathcal{T}_{M}}

\newcommand{\NBinx}{\mathcal{N}_{c}}
\newcommand{\NBiny}{\mathcal{N}_{r}}

\newcommand{\Bin}{\mathcal{B}}

\def\codeif{\mbox{\textbf{if\ }}}
\def\codefi{\mbox{\textbf{fi\ }}}
\def\codethen{\mbox{\textbf{then\ }}}
\def\codeforeach{\mbox{\textbf{for each\ }}}
\def\codefor{\mbox{\textbf{for\ }}}
\def\codedo{\mbox{\textbf{do\ }}}
\def\codewhile{\mbox{\textbf{while\ }}}
\def\codetrue{\mbox{\textbf{true\ }}}
\def\codefalse{\mbox{\textbf{false\ }}}
\def\codereturn{\mbox{\textbf{return\ }}}
\def\codebreak{\mbox{\textbf{break\ }}}
\def\codebool{\mbox{\textbf{bool\ }}}
\def\codeelse{\mbox{\textbf{else\ }}}
\def\codeswitch{\mbox{\textbf{switch\ }}}
\def\codecase{\mbox{\textbf{case\ }}}
\def\codeand{\mbox{\textbf{and\ }}}
\def\codeend{\mbox{\textbf{end\ }}}

\newcommand{\ind}{\quad}
\newcommand{\tallind}{\quad\quad}

\maketitle

\begin{abstract}
	A number of computer vision tasks exploit a succinct representation of the visual content in the form of sets of local features. Given an input image, feature extraction algorithms identify a set of keypoints and assign to each of them a description vector, based on the characteristics of the visual content surrounding the interest point. Several tasks might require local features to be extracted from a video sequence, on a frame-by-frame basis. Although temporal downsampling has been proven to be an effective solution for mobile augmented reality and visual search, high temporal resolution is a key requirement for time-critical applications such as object tracking, event recognition, pedestrian detection, surveillance. In recent years, more and more computationally efficient visual feature detectors and decriptors have been proposed. Nonetheless, such approaches are tailored to still images. In this paper we propose a fast keypoint detection algorithm for video sequences, that exploits the temporal coherence of the sequence of keypoints. According to the proposed method, each frame is preprocessed so as to identify the parts of the input frame for which keypoint detection and description need to be performed. Our experiments show that it is possible to achieve a reduction in computational time of up to 40\%, without significantly affecting the task accuracy.  
\end{abstract}

\begin{keywords}
Local features, keypoint detection, video.
\end{keywords}

\vspace{-1 mm}

\section{Introduction}
In recent years, ubiquitous computer vision applications are pervading our lives. Smartphones, self-driving terrestrial and aerial vehicles, Visual Sensor Networks (VSNs) are capable of acquiring visual data and performing complex analysis tasks. In particular, VSNs are expected to play a major role in the advent of the \emph{Internet-of-Things} paradigm. Such computer vision tasks usually exploit a concise yet effective representation of the acquired visual content, rather than being based on the pixel-level content. In this context, local features represent an effective solution that is being successfully exploited for a number of tasks such as content-based retrieval, object tracking, image registration, etc. Local feature extraction algorithms usually consist of two distinct components. First, a keypoint detector aims at identifying salient regions (e.g. corners, blobs) within a given image. Second, a descriptor assigns each identified keypoint a descriptor, in the form of a set of values, based on the local characteristics of the image patch surrounding such keypoint. Such information is further processed in order to extract a semantic representation of the acquired content, e.g., by identifying and tracking objects, recognizing faces, monitoring the environment and recognizing events. 

As regards visual feature extraction algorithms, SIFT~\cite{DBLP:journals:ijcv:Lowe04} is widely considered as the state-of-the-art for a large number of tasks. It consists in a keypoint detector based on the Difference-of-Gaussians (DoG) algorithm, and in a scale- and rotation-invariant real-valued descriptor, based on local intensity gradients. Besides, SURF~\cite{DBLP:conf:eccv:BayTG06} is partially inspired by SIFT and aims at achieving a similar level of accuracy at a lower computational cost. More recently, several low-complexity algorithms have been proposed, with the objective of alleviating the computational burden required by both traditional keypoint detectors and descriptors. For example, FAST~\cite{rosten_2005_tracking} and AGAST~\cite{mair2010_agast} are computationally efficient detectors capable of identifying stable corners. As for descriptors, binary-valued features are emerging as an efficient alternative to traditional real-valued features. BRIEF~\cite{DBLP:conf/eccv/CalonderLSF10}, BRISK~\cite{DBLP:conf/iccv/LeuteneggerCS11}, FREAK~\cite{DBLP:conf/cvpr/AlahiOV12} and BAMBOO~\cite{baroffio:2014:ICIP:BAMBOO} are instances of such category. For each identified keypoint, they compute a descriptor vector in the form of a sequence of binary values, each of which is obtained by comparing the (smoothed) intensities of a pair of pixels sampled around the keypoint. 
In some cases, ad-hoc software-based implementations are available for specific hardware architectures~\cite{baroffio:2014:ICIP:BRISKOLA}. 

Local feature detection in video sequences has been addressed in the past literature, with the goal of identifying keypoints that are stable across time. For example, Shi and Tomasi~\cite{Shi94goodfeatures} propose a widely adopted detector suitable for tracking applications. Zhang et. al propose a complex video-retrieval system based on color, shape and texture features extracted from the key-frames of a video~\cite{Zhang1997643}. More recently, Zha et al. propose a method to extract spatio-temporal features from video content~\cite{2009:Zha:ACCV}. Besides being a key to tasks such as object tracking, event identification and video calibration, temporally stable features improve the efficiency of coding architectures tailored to features extracted from video content~\cite{BaroffioRCTT:TIP, BaroffioRCTT:ICIP2013, Makar:2014:TIP}.
More recently, Girod et al.~\cite{journals/ijsc/MakarTCCG13} propose a feature detection and coding algorithm inspired by traditional motion estimation methods. Such algorithm selects a set of features corresponding to canonical image patches whose content is stable across frames, leading to a significant reduction of the transmission bitrate thanks to ad-hoc coding primitives. Although such algorithm represents a good solution for applications that require the efficient transmission of local features for further processing, it might not be the best in terms of computational complexity. Considering low-power devices, computationally intensive operations might significantly reduce the detection frame rate, possibly impairing performance of time-critical tasks or introducing undue delay. 
In this paper, we introduce a fast detection algorithm based on BRISK~\cite{DBLP:conf/iccv/LeuteneggerCS11} and tailored to the context of video sequences, aimed at reducing the computational complexity and thus enabling high frame rates, without significantly affecting performance in terms of accuracy.

The rest of this paper is organized as follows. Section~\ref{sec:BRISK} introduces the main concepts behind BRISK. Section~\ref{sec:algorithm} illustrates the proposed fast detection architecture. Section~\ref{sec:experiments} defines the experimental setup and presents results. Finally, conclusions are drawn in Section~\ref{sec:conclusions}.

\section{Binary Robust Invariant Scalable Keypoints (BRISK)}\label{sec:BRISK}

Leutenegger et al.~\cite{DBLP:conf/iccv/LeuteneggerCS11} propose the Binary Robust Invariant Scalable Keypoints (BRISK) algorithm as a computationally efficient alternative to traditional local feature detectors and descriptors. The algorithm consists in two main steps: i) a keypoint detector, that identifies salient points in a scale-space and ii) a keypoint descriptor, that assigns each keypoint a rotation- and scale- invariant binary descriptor. Each element of such descriptor is obtained by comparing the intensities of a given pair of pixels sampled within the neighborhood of the keypoint at hand. 

The BRISK detector is a scale-invariant version of the lightweight FAST~\cite{rosten_2005_tracking} corner detector, based on the Accelerated Segment Test (AST). Such a test classifies a candidate point $p$ (with intensity $I_p$) as a keypoint if $n$ contiguous pixels in the Bresenham circle of radius 3 around $p$ are all brighter than $I_p + t$, or all darker than $I_p - t$, with $t$ a predefined threshold. Thus, the highest the threshold, the lowest the number of keypoints which are detected and vice-versa. 

Scale-invariance is achieved in BRISK by building a scale-space pyramid consisting of a pre-determined number of octaves and intra-octaves, obtained by progressively downsampling the original image. 
The FAST detector is applied separately to each layer of the scale-space pyramid, in order to identify potential regions of interest having different sizes. Then, non-maxima suppression is applied in a 3x3 scale-space neighborhood, retaining only features corresponding to local maxima. Finally, a three-step interpolation process is applied in order to refine the correct position of the keypoint with sub-pixel and sub-scale precision.
\vspace{-2 mm}

\section{Fast video feature extraction}\label{sec:algorithm}
Let $\Image_n$ denote the $n$-th frame of a video sequence of size $N_x \times N_y$, 
which is processed to extract a set of local features $\Descr_n$. First, a keypoint detector is applied to identify a set of interest points. Then, a descriptor is applied on the (rotated) patches surrounding each keypoint. Hence, each element of $\descr_{n,i} \in \Descr_n$ is a visual feature, which consists of two components: i) a 4-dimensional vector $\coord_{n,i} = [x, y, \scale, \orient]^T$, indicating the position $(x,y)$, the scale $\scale$ of the detected keypoint, and the orientation angle $\theta$ of the image patch; ii) a $\Dim$-dimensional \emph{binary} vector $\dv_{n,i} \in \{0,1\}^\Dim$, which represents the descriptor associated to the keypoint $\coord_{n,i}$.

Traditionally, local feature extraction algorithms have been designed to efficiently extract and describe salient points within a single frame. Considering video sequences, a straightforward approach consists in applying a feature extraction algorithm separately to each frame of the video sequence at hand. 
However, such a method is inefficient from a computational point of view, as the temporal redundancy between contiguous frame is not taken into consideration. The main idea behind our approach is to apply a keypoint detection algorithm only on some regions of each frame. To this end, for each frame $\Image_n$, a binary \emph{Detection Mask} $\Mask_n \in \{0, 1\}^{N_x \times N_y}$ having the same size of the input image is computed, exploiting the information extracted from previous frames. Such mask defines the regions of the frame where a keypoint detector has to be applied. That is, considering an image pixel $\Image_n(x,y)$, a keypoint detector is applied to such a pixel if the corresponding mask element $\Mask_n(x,y)$ is equal to 1. Furthermore, we assume that if a region of the $n$-th frame is not subject to keypoint detection , the keypoints that are present in such an area in the previous frame, i.e. $\Image_{n-1}$, are still valid. Hence, such keypoints are propagated to the current set of features. That is,

\begin{equation}
	\Descr_n = \{\descr_{n,i} : \Mask_n(\coord_{n,i}) = 1 \;\cup\; \descr_{n-1, j} : \Mask_n(\coord_{n-1, j}) = 0\}
\end{equation} 

Note that the algorithm used to compute the \emph{Detection Mask} needs to be computationally efficient, so that the savings achievable by skipping detection in some parts of the frame are not offset by this extra cost. In the following, two efficient algorithms for obtaining a \emph{Detection Mask} are proposed: \emph{Intensity Difference Detection Mask} and \emph{Keypoint Binning Detection Mask}.

\vspace{-3 mm}

\subsection{Intensity Difference Detection Mask}\label{sec:IDDM}

The key tenet is to apply the detector only to 
those regions that change significantly across the frames of the video. In order to identify such regions and build the \emph{Detection Mask}, we exploit the scale-space pyramid built by the BRISK detector, thus incurring in no extra cost. 
Considering frame $\Image_n$ and $\NOct$ detection octaves, pyramid layers $\Pyr_{n,o}, o = 1, \dots, \NOct$ are obtained by progressively smoothing and half-sampling the original image, as illustrated in Section~\ref{sec:BRISK}. Then, considering two contiguous frames $\Image_{n-1}$ and $\Image_{n}$ and octave $o$, a subsampled version of the \emph{Detection Mask} is obtained as follows:

\begin{equation}
	\Mask'_{n,o}(k, l) =
		\begin{cases} 1 &\mbox{if } \lvert \Pyr_{n, o}(k,l) - \Pyr_{n-1, o}(k,l) \rvert \le \ThresID \\ 
			0 & \mbox{if } \lvert \Pyr_{n, o}(k,l) - \Pyr_{n-1, o}(k,l) \rvert > \ThresID,
		\end{cases}
\end{equation}

where $\ThresID$ is an arbitrarily chosen threshold and $(k,l)$ the coordinates of the pixels in
the intermediate representation $\Mask'_{n,o}$. 
Finally, the intermediate representation $\Mask'_{n,o}$ resulting from the previous operation needs to be upsampled in order to obtain the final mask  $\Mask_n \in \{0, 1\}^{N_x \times N_y}$. Masks can then be applied to detection in different fashions: i) exploiting the mask obtained resorting to each scale-space layer $o = 1, \dots, \NOct$ in order to detect keypoint at the corresponding layer $o$; ii) use a single detection mask for all the scale-space layers. 

\vspace{-3 mm}

\subsection{Keypoint Binning Detection Mask}\label{sec:KBDM}

Considering two contiguous frames of a video sequence, the amount of features identified in a given area are often correlated~\cite{Eriksson:2014:ICASSP}. To exploit such information, 
the detector is applied to a region of the input image only if the number of features extracted in the co-located region in the previous frame is greater than a threshold. Specifically, in order to obtain a \emph{Detection Mask} for the $n-$th frame, a spatial binning process is applied to the features extracted from frame $\Image_{n-1}$. To this end, we define a grid consisting of $\NBiny \times \NBinx$ spatial bins $\Bin_{i, j}, i = 0, \dots, \NBiny, j = 0, \dots, \NBinx$. Thus, each bin refers to a rectangular area of $S_x \times S_y$ pixels, where $S_x = \nicefrac[]{N_x}{\NBinx}$ and $S_y = \nicefrac[]{N_y}{\NBiny}$. Then, a two-dimensional spatial histogram of keypoints is created by assigning each feature to the corresponding bin as follows:

\begin{equation}
	\Mask''_{n}(k, l) = \lvert \descr_{n-1, i} \in \Descr_{n-1} \rvert : \lfloor \nicefrac{x_{n-1, i}}{S_x}\rfloor = k , \lfloor \nicefrac{y_{n-1, i}}{S_y}\rfloor = l,
\end{equation}

where $(x_{n-1, i}, y_{n-1, i})$ represents the location of feature $\descr_{n-1, i}$ and $|\cdot|$ the number of elements in a set.
Then, a binary subsampled version of the \emph{Detection Mask} is obtained by thresholding such histogram, employing a tunable threshold $\ThresKH$:

\begin{equation}
	\Mask'_{n}(k, l) =
		\begin{cases} 1 &\mbox{if } \Mask''_{n}(k, l) \ge \ThresKH \\ 
			0 & \mbox{if } \Mask''_{n}(k, l) < \ThresKH,
		\end{cases}
\end{equation}

Finally, the \emph{Detection Mask} $\Mask_n$ having size $N_x \times N_y$ pixels is obtained by upsampling the intermediate representation $\Mask'_{n}$. Such a detection mask is applied to all scale-space octaves.

\vspace{-2 mm}

\section{Experiments}\label{sec:experiments}
\textbf{Dataset: }
We evaluated the proposed algorithms with respect to three different test scenarios. First, we exploited the \emph{Stanford MAR dataset}~\cite{Makar:2014:TIP}, containing the four VGA size, 200 frames long video sequences \emph{Alicia Keys}, \emph{Fogelberg}, \emph{Anne Murray} and \emph{Reba}. Each sequence contains a CD cover recorded with a hand-held mobile phone, under different imaging conditions such as illumination, zoom, perspective, rotation, glare, etc. Furthermore, for each sequence, the dataset contains the ground truth information, in the form of a still image of the corresponding CD cover, having a size of $500 \times 500$ pixels. 

As a second test, we evaluated the approaches resorting to the \emph{Rome Landmark Dataset}. Such dataset includes a set of 10 query video sequences, each capturing a different landmark in the city of Rome with a camera embedded in a mobile device~\cite{romelandmark}. The frame rate of such sequences is equal to 24fps, whereas the resolution ranges from 480x360 pixels (4:3) to 640x360 pixels (16:9). The first 50 frames of each video were used as query. On average, each query video corresponds to 9 relevant images representing the same physical object under different conditions and with heterogeneous qualities and resolutions. Then, distractor images randomly sampled from the \emph{MIRFLICKR-1M} dataset~\cite{huiskes08}, so that the final database contains 10k images.

Finally, we tested our method on the \emph{Stanford MAR multiple object} video set~\cite{Makar:2014:TIP}. Such a set is made up of 4 video sequences, each consisting of 200 frames at 640x480 resolution. Each video is recorded with a handheld camera and portrays three different objects, one at a time.

\textbf{Methods: }
We tested the two detection methods presented in Section~\ref{sec:algorithm}, that is, \emph{Intensity Difference Detecion Mask} and \emph{Keypoint Binning Detection Mask}. In both cases, we employed the original BRISK implementation from the authors\footnote{http://www.asl.ethz.ch/people/lestefan/personal/BRISK}, setting the number of octaves to 4 and the detection threshold to 55 and 70 for the \emph{Stanford MAR dataset} and the \emph{Rome landmark dataset}, respectively.
As regards \emph{Intensity Difference Detection Mask}, we built the mask testing several different configurations. 
We tested our algorithm with the 4 layers corresponding to each scale-space octaves.  
Since the performance was similar when using different layers, we resorted to the top-layer, i.e., the one with the lowest spatial resolution and processing cost. 
%
Both \emph{Intensity Difference Detection Mask} and \emph{Keypoint Binning Detection Mask} require a threshold to be set in order to obtain the final detection mask. We tested several different configurations, each representing a tradeoff between computational efficiency and task accuracy. 

We compared our algorithms with a \emph{Temporally Coherent Detector} based on non-canonical patch matching~\cite{Makar:2014:TIP}, which also exploits temporal redundancy in the detected keypoints.
%
Such algorithm aims at propagating stable keypoints across frames, exploiting a pixel-level representation of local features. In details, a traditional keypoint detector is applied to the first frame of a Group Of Pictures of size $\Delta$. Given an identified keypoint, a non-canonical square image patch is extracted from the neighborhood of such a point. Then, considering the following frame, we searched for a matching patch in a window surrounding such a keypoint. Two patches are assumed to be a match if the Sum of Absolute Differences (SAD) between their pixels is below a given threshold $\ThresBM$. Finally, keypoints for which a match is found are propagated to the next frame, and their position is determined by the aforementioned block matching procedure. 
In our tests, according to the prescriptions of~\cite{Makar:2014:TIP}, we employed patches of $16 \times 16$ pixels and we set $\Delta = 10$ and $\ThresBM = 1800$. Furthermore, to make the procedure faster, we implemented a coarse-to-fine matching algorithm, where the first step consists in a spiral search algorithm with a precision of $2$ in a search window of $24 \times 24$ pixels, whereas the second step in a spiral search algorithm with quarter-pixel precision in a search window of $1.75 \times 1.75$ pixels. Finally, to further speed-up the process, we set an early termination SAD threshold $\ThresET = 1000$. 
This detector was originally proposed with the goal of maximizing coding efficiency, when patches around the detected keypoints need to be compressed and transmitted. To this end, this method can also adopt more sophisticated matching strategies, e.g., based on affine warping. However, in this paper we consider an implementation based on block matching to minimize the computational complexity.

\textbf{Evaluation methods and measures: }In the case of the \emph{Stanford MAR dataset}, for a given video sequence, we extracted a set of features for each frame. 
Then, the set of features extracted from a frame is matched with the ones extracted from the ground truth frame. A radius match algorithm is used, where the matching threshold is set to $\ThresRM = 0.18*512 \simeq 102$. Finally, geometric coherence of matches is enforced resorting to the RANSAC algorithm. Finally, the number of \emph{Matches-Post-Ransac} (MPR) is employed as the accuracy measure. 

\begin{figure}[t]
	\centering
	\includegraphics[trim=0cm 0cm 0cm 0cm, clip=true,width=0.43\textwidth]{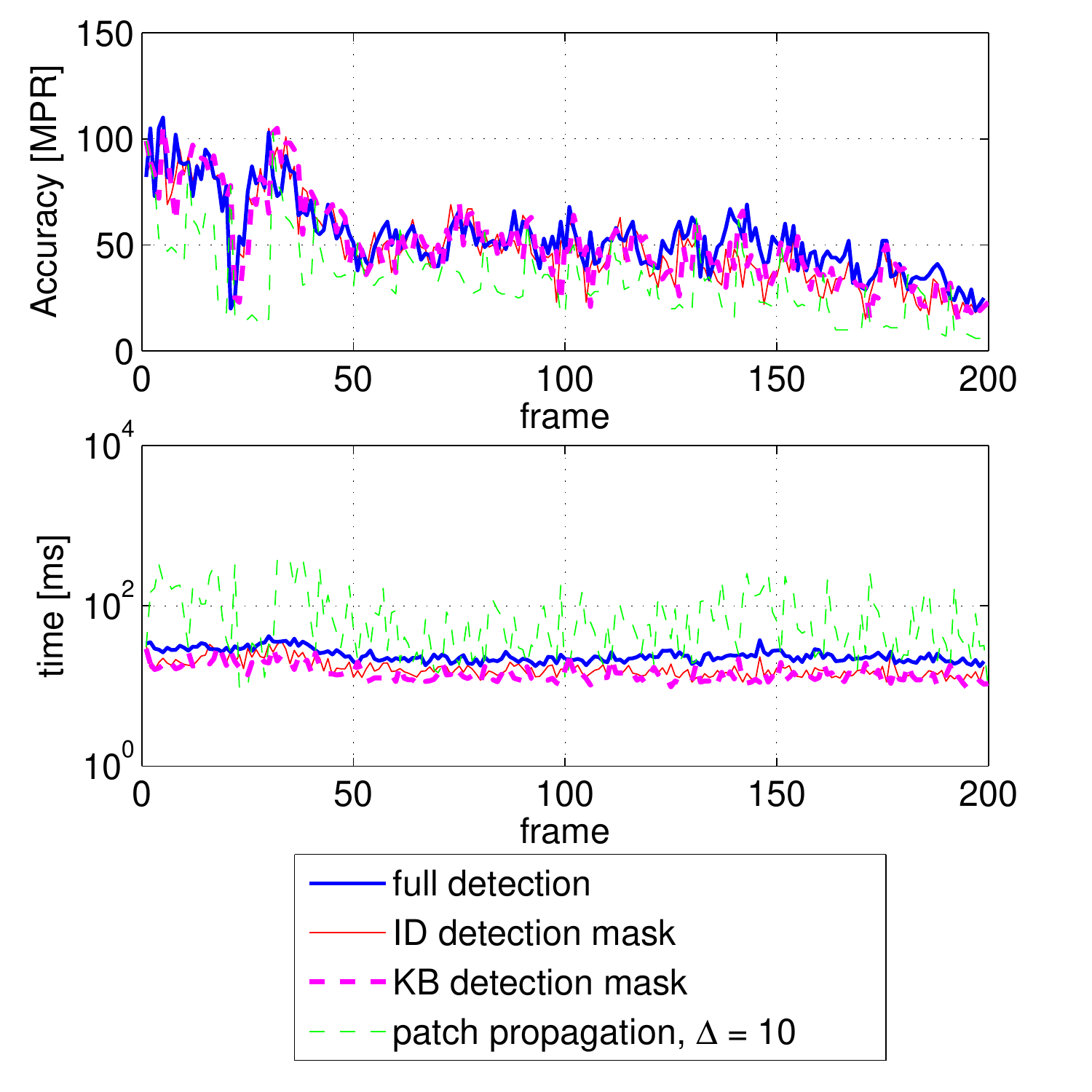}
	\caption{Accuracy, measured as the number of matches post-RANSAC (MPR), and computational time for each frame of the \emph{Alicia Keys} test sequence.}
	\label{fig:MPR_over_time_Alicia}
	\vspace{-3 mm}
\end{figure}

In the case of the \emph{Rome Landmark dataset}, the accuracy of the task was evaluated according to the \emph{Mean Average Precision} (MAP). Given an input query sequence $q$, for each frame $\Image_{q,n}$ it is possible to define the \emph{Average Precision} as 
\begin{equation}
	AP_{q,n} = \frac{\sum_{k=1}^Z P_{q,n}(k)r_{q,n}(k)}{R_{q,n}},
\end{equation}
where $P_{q,n}(k)$ is the precision (i.e., the fraction of relevant documents retrieved) considering the top-$k$ results in the ranked list of database images; $r_{q,n}(k)$ is an indicator function, which is equal to 1 if the item at rank $k$ is relevant for the query, and zero otherwise; $R_{q,n}$ is the total number of relevant document for frame $\Image_{q,n}$ of the query sequence $q$ and $Z$ is the total number of documents in the list.
The overall accuracy for the query sequence $q$ is evaluated according to

\begin{equation}
	AP_q = \frac{\sum_{n = 1}^N AP_{q,n}}{N}, 
\end{equation}

where $N$ is the total number of frames of the query video $q$. 

Finally, the \emph{Mean Average Precision} is obtained as
\begin{equation}\label{eq:MAP}
MAP = \frac{\sum_{q = 1}^Q AP_q}{Q}, 
\end{equation}
that is, the mean of the $MAP_q$ measure over all the query sequences.

In the case of the \emph{Stanford MAR multiple object} video set, the accuracy is measured according to a combined detection and tracking precision metric. In particular, for each frame, the goal is to correctly detect the portrayed database object and to identify its position within the frame. Each frame of the video sequences is matched against all the database object. Radius match and geometric verification steps are performed as in the case of~\emph{Stanford MAR dataset} scenario. The matching object is the one with the highest number of matches-post-RANSAC. The bounding box for the identified object is obtained by projecting the database object corners according to the homography computed with the RANSAC algorithm at the previous step. Each frame is deemed as correct if the correct object is detected, and if the estimated position is consistent with the ground-truth information. As to the latter, the estimated object position is deemed correct if the displacement between the estimated centroid and the ground truth one is lower than a threshold. We set the value of such a threshold to 10 pixels. 

We evaluated the complexity of the feature extraction methods by means of the required CPU time. We performed our tests on a laptop equipped with a 2.5GHz Intel Core i5 processor and 10 GB of RAM. 

\begin{figure}[t]
	\centering
	\includegraphics[trim=0.5cm 0cm 0cm 0cm, clip=true,width=0.40\textwidth]{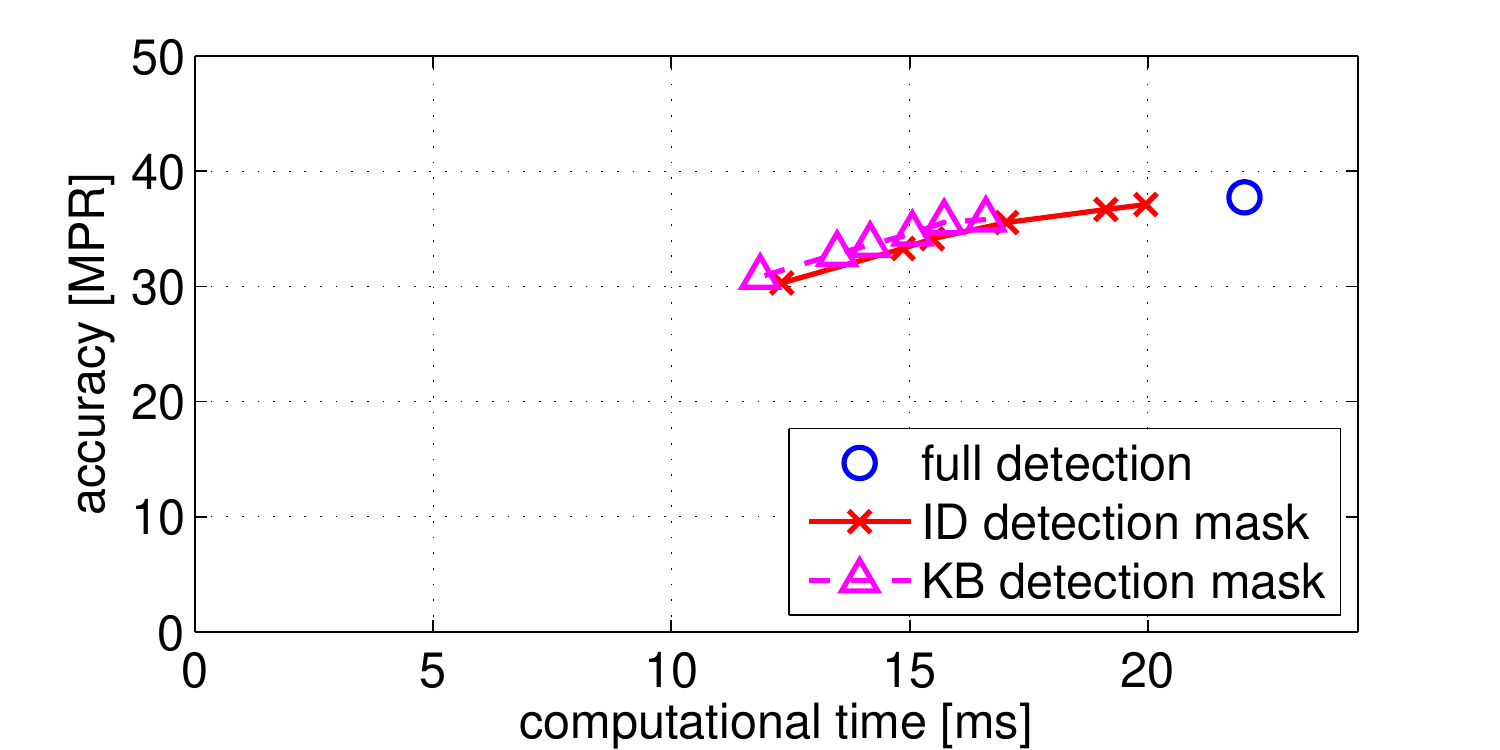}
	\caption{Energy-Accuracy curves for \emph{Stanford MAR} dataset.}
	\label{fig:energyaccuracy} 
	\vspace{-3 mm}
\end{figure}

\textbf{Results: } As an illustrative example, Figure~\ref{fig:MPR_over_time_Alicia} shows the results obtained for the \emph{Alicia Keys} sequence. The charts also report the results obtained when detection is performed independently on a frame-by-frame basis (full detection) to serve as a reference. 
We observe that the method using the \emph{Intensity Difference Detection Mask} (threshold 20) achieves an accuracy level similar to that of full detection ($MPR = 55$ vs. $56$), at a reduced computational time (20.5 ms vs. 24.5 ms). As for \emph{Temporally Coherent Detector}, it leads to a significant loss in terms of accuracy ($MPR = 41$), while being quite computationally intensive (72 ms on average). While accuracy could be further improved by resorting to matching based on affine warping, this would further increase its complexity. This confirms the fact that this detector was originally designed with the goal of maximizing coding efficiency rather than computational cost. Since this is confirmed also on other test sequences, we do not report additional results for this detector. 

It is interesting to observe the energy-accuracy trade-off that can be achieved by varying the threshold used by the algorithms based on detection masks. To this end,  Figure~\ref{fig:energyaccuracy} compares the performance of \emph{Intensity Difference Detection Mask} and \emph{Keypoint Binning Detection Mask} with that of full detection, averaging the results on the \emph{Stanford MAR dataset}. The two methods based on a detection mask performs on a par, reducing the required computational time by 30\% while losing as few as 4 matches. 

\begin{figure}[t]
	\centering
	\includegraphics[trim=0cm 0cm 0cm 0cm, clip=true,width=0.40\textwidth]{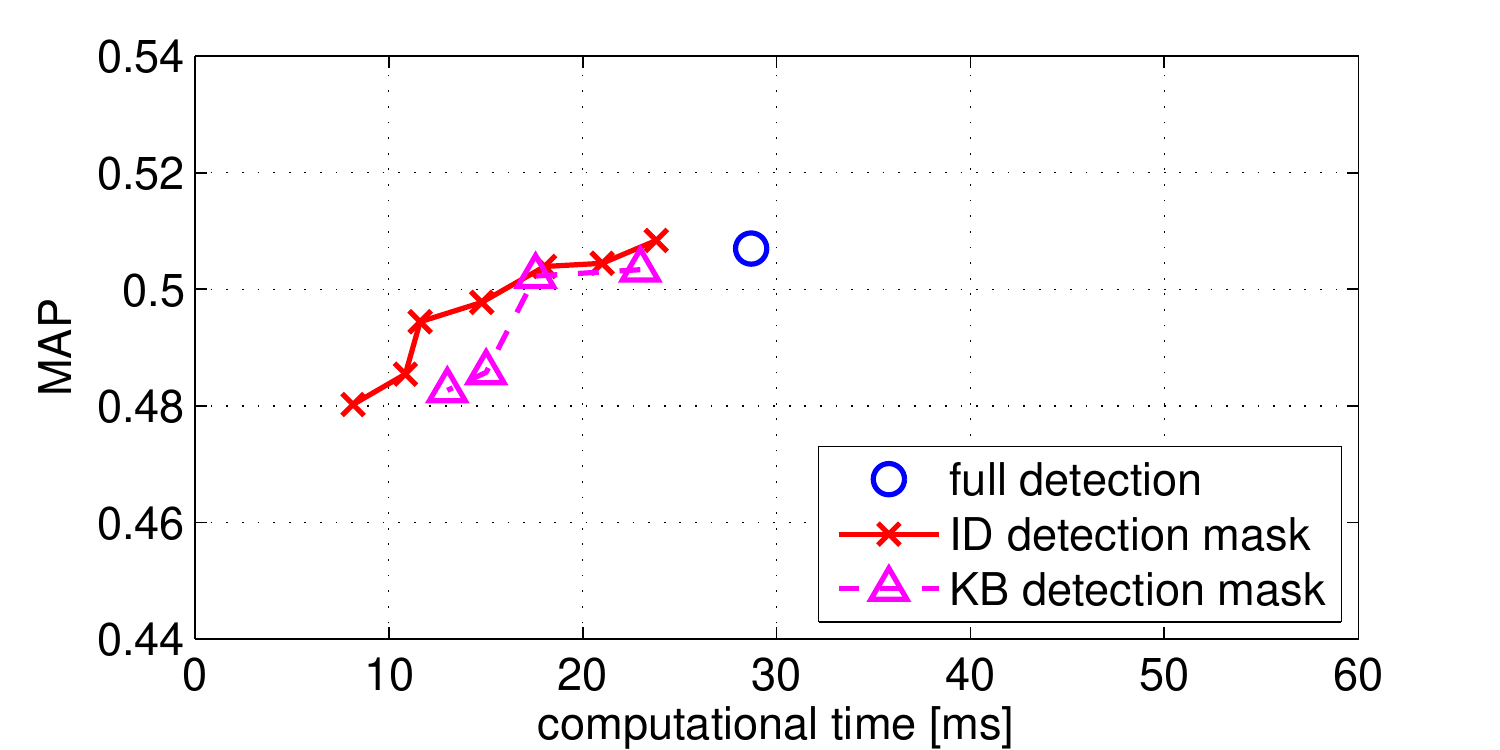}
	\caption{Energy-Accuracy curve for the \emph{Rome Landmark} dataset, when using the \emph{Intensity Difference Detection Mask} in order to reduce the detection area and with different values for the thresholding parameter. The computational time for each frame can be reduced from 28ms to 18ms, without significantly affecting the accuracy of the task.}
	\label{fig:RomeLandmarkMAP} 
	\vspace{-1 mm}
\end{figure}

Furthermore, we tested our approach based on a \emph{Detection Mask} on the \emph{Rome Landmark Dataset}. 
Figure~\ref{fig:RomeLandmarkMAP} compares the results of \emph{Intensity Difference Detection Mask} with that of full detection, showing that computational time can be reduced by about 35\% without affecting task accuracy. Furthermore, the feature extraction process can be speeded un by 3 times at the cost of 0.03\% lower \emph{Mean Average Precision}.

\begin{figure}[t]
	\centering
	\includegraphics[trim=0cm 0cm 0cm 0cm, clip=true,width=0.40\textwidth]{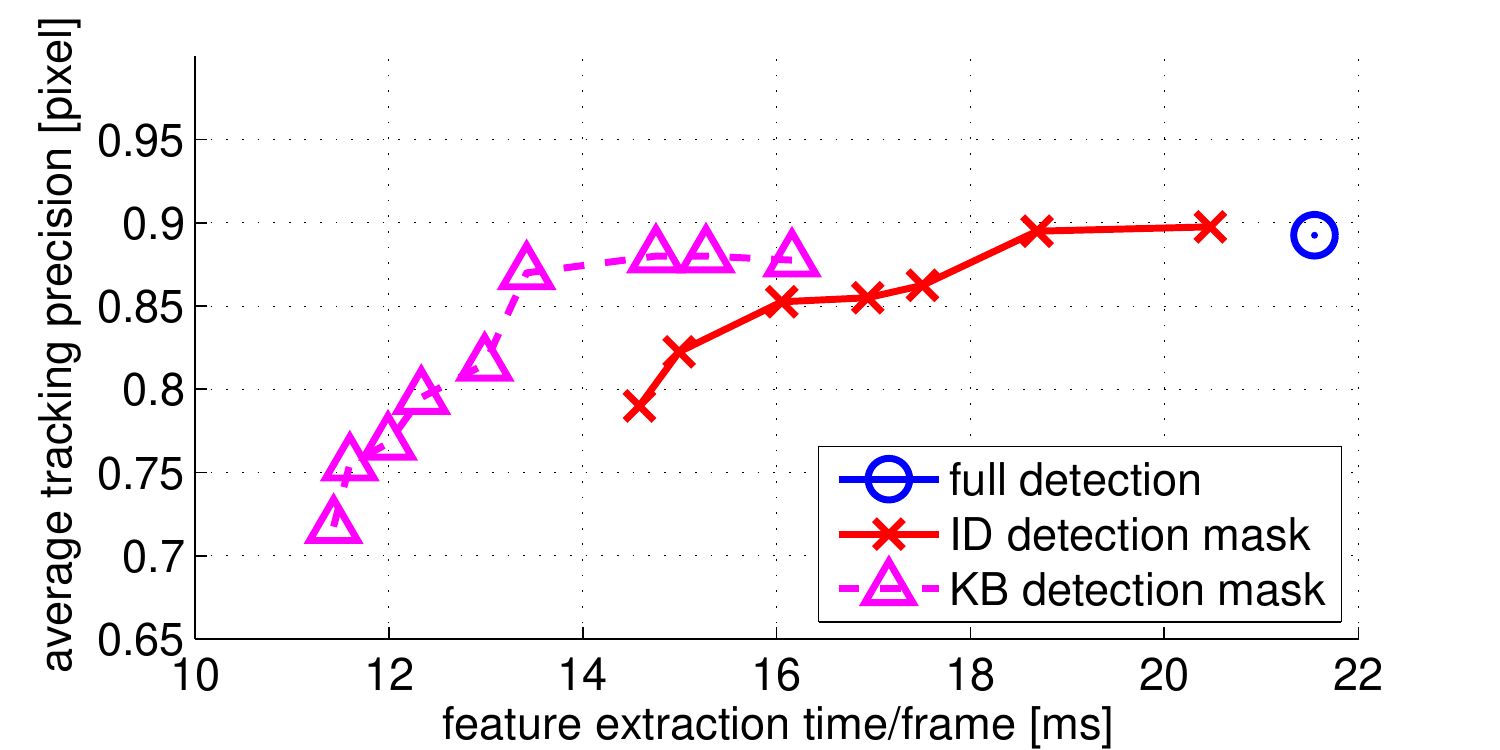}
	\caption{Energy-Accuracy curves for the \emph{Stanford MAR} multiple object sequences.}
	\label{fig:StanfordMultiple} 
	\vspace{-4 mm}
\end{figure}

Finally, the results of our fast detection algorithms on the \emph{Stanford MAR multiple object} video set are reported in Figure~\ref{fig:StanfordMultiple}. The computational time can be be reduced up to 40\% without significantly impairing object detection and tracking performance.

\vspace{-2 mm}

\section{Conclusions}\label{sec:conclusions}

In this paper we presented a method for fast keypoint detection in video sequences based on \emph{Detection Masks}. Results show that the proposed approach allows for a reduction in terms of computational complexity of up to 35\% without significantly impair task performance. 
In our future investigation we plan to further improve the \emph{Detection Mask} building process, by introducing more sophisticated yet computationally efficient solutions. 


\balance

\bibliographystyle{IEEEbib}
\bibliography{IEEEabrv,string-defs,bibfile}

\end{document}